%% file: main.tex
\documentclass{article}
\usepackage{subfiles}
\usepackage{ijcai25}

\usepackage{times}
\usepackage{soul}
\usepackage{url}
\usepackage[hidelinks]{hyperref}
\usepackage[utf8]{inputenc}
\usepackage[small]{caption}
\usepackage{graphicx}
\usepackage{amsmath}
\usepackage{amsthm}
\usepackage{booktabs}
\usepackage{algorithm}
\usepackage{algorithmic}
\usepackage[switch]{lineno}
\usepackage{amssymb}
\usepackage{amsfonts}
\usepackage{multirow}
\usepackage{color}
\usepackage{hyperref}
\usepackage{stfloats}

\title{FedSaaS: Class-Consistency Federated Semantic Segmentation via Global Prototype Supervision and Local Adversarial Harmonization}

\author{
Xiaoyang Yu$^{1,2}$\and
Xiaoming Wu$^{1,2}$\and
Xin Wang$^{1,2}$\and
Dongrun Li$^{1,2}$\and
Ming Yang$^{1,2}$\and
Peng Cheng$^3$\\
\affiliations
$^1$Key Lab. of Computing Power Network and Information Security, Ministry of Education, Shandong Computer Science Center, Qilu University of Technology (Shandong Academy of Sciences), Jinan, China\\
$^2$Shandong Provincial Key Laboratory of Industrial Network and Information System Security, Shandong Fundamental Research Center for Computer Science, Jinan, China\\
$^3$College of Control Science and Engineering, Zhejiang University, Hangzhou, China\\
\emails
b1043124010@stu.qlu.edu.cn,
wuxm@sdas.org,
xinwang@qlu.edu.cn
}

\begin{document}

\maketitle
\begin{abstract}
Federated semantic segmentation enables pixel-level classification in images through collaborative learning while maintaining data privacy. However, existing research commonly overlooks the fine-grained class relationships within the semantic space when addressing heterogeneous problems, particularly domain shift. This oversight results in ambiguities between class representation. To overcome this challenge, we propose a novel federated segmentation framework that strikes class consistency, termed FedSaaS. Specifically, we introduce class exemplars as a criterion for both local- and global-level class representations. On the server side, the uploaded class exemplars are leveraged to model class prototypes, which supervise global branch of clients, ensuring alignment with global-level representation. On the client side, we incorporate an adversarial mechanism to harmonize contributions of global and local branches, leading to consistent output. Moreover, multilevel contrastive losses are employed on both sides to enforce consistency between two-level representations in the same semantic space. Extensive experiments on several driving scene segmentation datasets demonstrate that our framework outperforms state-of-the-art methods, significantly improving average segmentation accuracy and effectively addressing the class-consistency representation problem.
    
\end{abstract}

\section{Introduction}

Semantic segmentation plays a pivotal role in various fields, such as autonomous driving \cite{feng2020deep}, medical diagnosis \cite{qureshi2023medical}, and remote sensing \cite{yuan2021review}, where precise pixel-level classification is critical. While deep learning has significantly improved segmentation accuracy through the use of large datasets, the need for extensive labeled data is often hindered by concerns surrounding data privacy and security. Federated learning (FL) addresses this challenge by facilitating collaborative, decentralized model training while preserving data privacy \cite{mcmahan2017communication}. As such, FL represents a promising paradigm for enabling cross-region semantic segmentation. 

\begin{figure}[!t]
\centering
\includegraphics [width=\columnwidth] {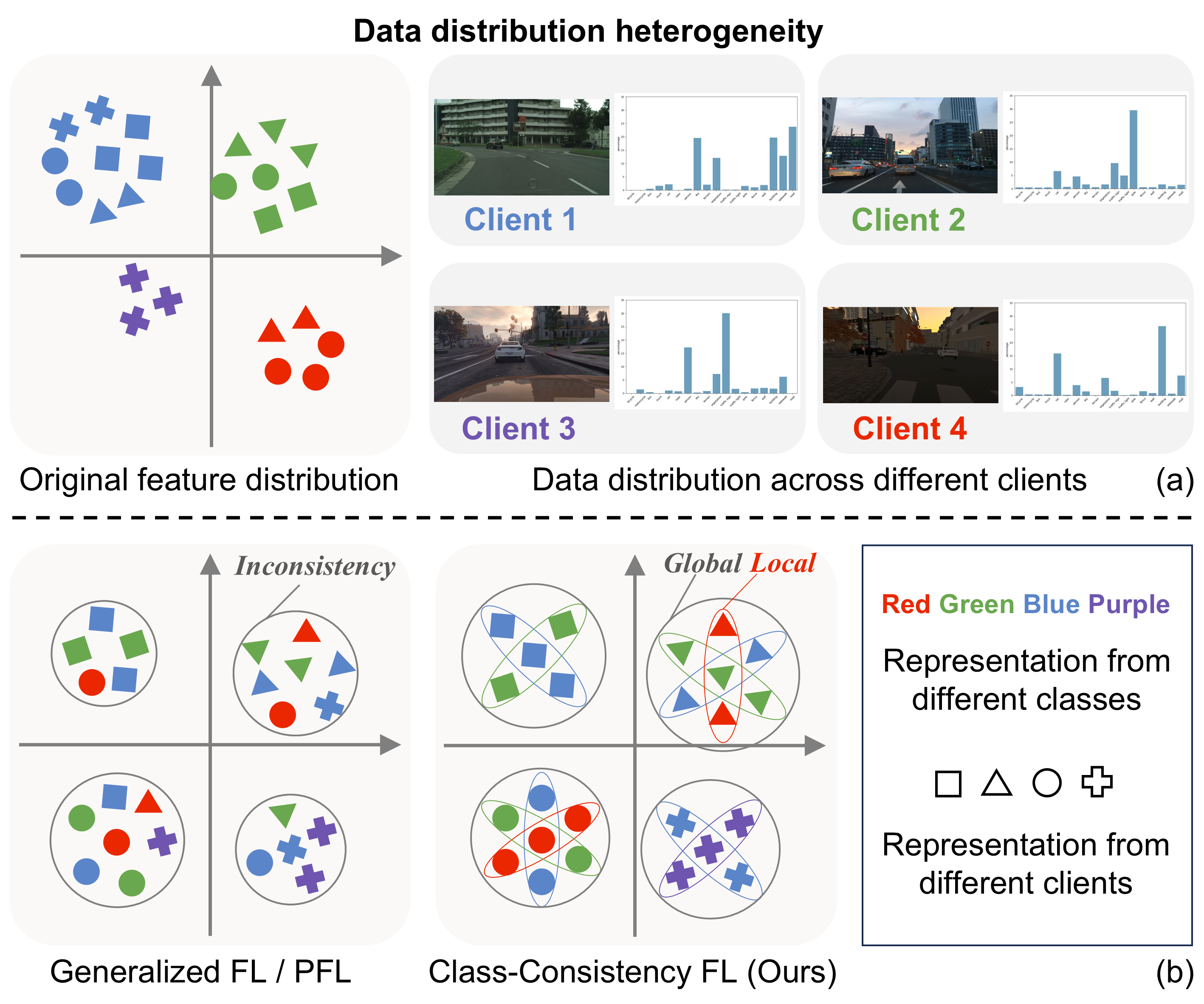}
\caption{(a) Data from different regions exhibits severe domain shift and label skew. In the semantic space, the same class often shows significantly different distributions across clients. (b) Existing FL methods primarily focus on generalization or personalization (left) but often ignore class-level alignment. Class-consistency FL (right) ensures consistent representation of the same class by aligning and constraining both global- and local-level representations.}
\label{fig1}
\end{figure}
    
However, federated semantic segmentation continues to face significant challenges due to the heterogeneity of client data. In real-world applications, data from different clients often exhibit substantial variation, stemming from factors like sensor biases, environmental changes, and user preferences. This heterogeneity gives rise to discrepancies in the learned representations, which complicates the model's ability to generalize across clients \cite{huang2023rethinking}. This issue becomes particularly prominent in FL when clients share the same objects but exhibit distinct feature distributions, a phenomenon known as domain shift. As illustrated in Figure~\ref{fig1}(a), driving images from different domains exemplify this challenge: data from clients~1 and~2 are derived from real street scenes in different regions, while clients~3 and~4 consist of synthetically generated data. Domain shift factors can cause the same objects to appear visually distinct, leading to misaligned representation distributions in the semantic space and impeding accurate class identification.

Existing research often overlooks the fine-grained class relationships within the semantic space when addressing domain shift factors. In general FL methods, such as \cite{wang2024fedsiam,xu2024fblg,collins2021exploiting}, clients focus on extracting effective local-level representations from their specific domains, while the server aims to capture global-level representations shared across multiple domains. Since these two types of representations correspond to different scales of data understanding, inconsistencies may arise between local and global representations in the semantic space. This issue is particularly pronounced in federated semantic segmentation tasks, where both local- and global-level representations are crucial for capturing fine-grained class-level relationships, both intra-class and inter-class. Existing solutions \cite{ma2024fedst,miao2023fedseg,kou2024pfedlvm} often employ style transfer and contrastive learning to enhance generalization or adapt to local characteristics. However, these approaches fail to address the inconsistency problem of class representations between local and global semantic spaces. As shown in Figure~\ref{fig1}(b), such inconsistencies can lead to semantic mismatches, where the same class is represented differently across domains. This leads to divergent semantic interpretations, ultimately impairing the model’s ability to generalize effectively. To address the challenge of class inconsistency, this paper explores methods for aligning and constraining both local and global class representations within the semantic space. 

In this work, we propose a class-consistency \textbf{Fed}erated semantic \textbf{S}egment\textbf{a}tion approach via glob\textbf{a}l prototype \textbf{S}upervision and local adversarial harmonization, termed \textbf{FedSaaS}. To measure class representations at both global and local levels in the semantic space, we introduce class exemplars as a criterion, inspired by mask average pooling \cite{siam2019amp}. FedSaaS leverages class exemplars on the server side to train the global model and generate class prototypes that supervise the global branch of the client model, ensuring alignment with global-level class representations. On the client side, we integrate an adversarial mechanism to harmonize the contributions of the local and global branches, thereby achieving consistent outputs. Furthermore, multilevel contrastive losses based on class exemplars are employed on both the client and server sides to enforce consistency within the same semantic space. We evaluate FedSaaS on five autonomous driving scene datasets, constructing datasets with varying levels of heterogeneity (slight and severe). Experimental results demonstrate that FedSaaS outperforms state-of-the-art methods across different degrees of domain shift. 

The contributions of this paper are highlighted as follows:

\begin{itemize}
    \item We propose FedSaaS, a federated semantic segmentation framework that introduces class exemplars to achieve consistency in class representations at both global and local levels within the semantic space. 
    \item We supervise the alignment of local class representations by modeling class prototypes and integrate an adversarial mechanism within the client to harmonize the contributions of the global and local branches, thereby ensuring consistent outputs. Multilevel contrastive losses are introduced to further enhance the consistency between the two-level representations.
    \item Experiments on driving scene datasets demonstrate the superior performance of FedSaaS. Ablation studies and empirical analyses further validate its effectiveness in achieving consistency, improving segmentation precision, and enhancing communication efficiency.
\end{itemize}

\section{Related Work}

\textbf{Semantic Segmentation}. It is a task that assigns each pixel in an image to a predefined category. State-of-the-art methods predominantly employ encoder-decoder architectures based on various network models, including convolutional neural networks \cite{long2015fully,ronneberger2015u,zhao2017pyramid,chen2017deeplab}, vision transformers \cite{dosovitskiy2020image,liu2021swin}, diffusion models \cite{tian2024diffuse,amit2021segdiff}, and Mamba \cite{xing2024segmamba}. These methods typically rely on large-scale labeled datasets and centralized training, achieving high segmentation accuracy on public benchmarks. However, the centralized training paradigm raises significant concerns related to data privacy and accessibility. In response,  recent research has begun to explore distributed training frameworks as a promising alternative in such contexts.

\begin{figure*}[!t]
\centering
\includegraphics [width=\textwidth] {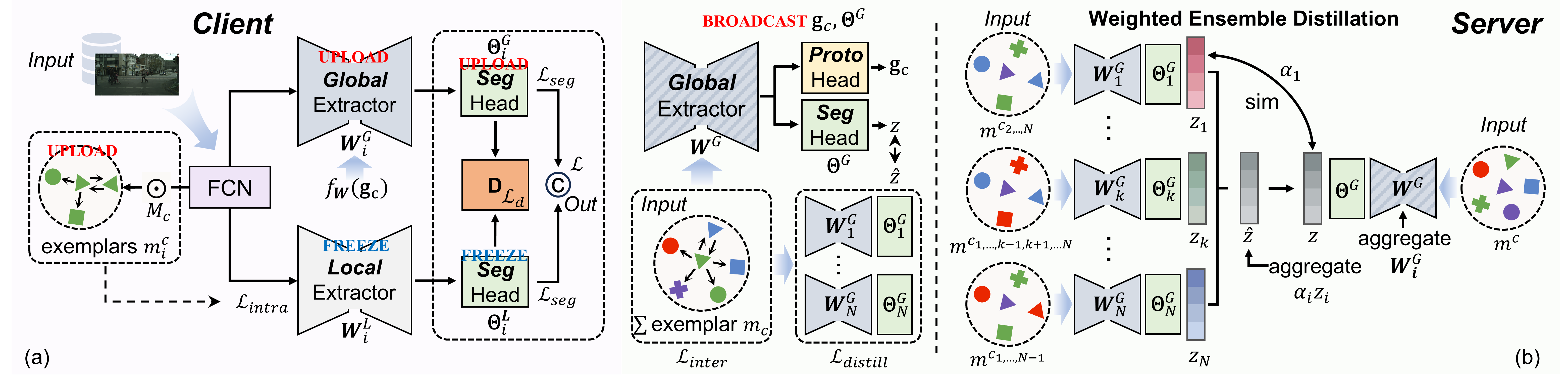}
\caption{Overview of the FedSaaS Framework. (a) Client-side details. (b) Server-side and weighted integrated distillation details.}
\label{fig2}
\end{figure*}

\textbf{Federated Learning}. FL is a suitable paradigm for meeting the distributed training requirements in semantic segmentation tasks. 
This approach has made significant progress in addressing generalization challenges such as label shift and domain shift \cite{wang2024fedsiam,xu2024fblg}. Personalized federated learning (PFL) further extends this paradigm by accounting for the specific needs of each client, adapting the global model to better align with the data distribution of individual clients \cite{collins2021exploiting}. Nevertheless, unlike traditional FL tasks, semantic segmentation introduces unique challenges due to its requirement for precise pixel-level classification. This necessitates the retention of fine-grained spatial information at the local level, which complicates the direct application of existing general FL methods. These challenges are particularly pronounced in scenarios with severe data heterogeneity. 

\textbf{Federated Semantic Segmentation}. The problem of FL-based semantic segmentation was first studied by \cite{michieli2021prototype}. Existing methods can be broadly categorized into two types: local personalization and global generalization. The first type primarily focuses on PFL frameworks, aiming to adapt local models to the specific characteristics of each client \cite{tan2022towards}. This approach has been widely applied in fields such as medical image analysis and autonomous driving \cite{xie2024pflfe,wang2023feddp,kou2024pfedlvm}. These works typically fuse various forms of local features learned by clients with shared features on the server to align with the distribution of local data. However, the fusion process often lacks appropriate constraints, which can result in an overemphasis on either global or local features, leading to an imbalance between local and global outputs. To address this limitation, we introduce an adversarial mechanism within each client, which promotes mutual learning between local and global branches, thereby facilitating a balanced contribution from both. The second category, global generalization, aims to enhance generalization by learning shared knowledge across domains, with a particular focus on addressing style heterogeneity within each domain \cite{ma2024fedst,fantauzzo2022feddrive}. Commonly adopted techniques include knowledge distillation \cite{wang2023dafkd}, ensemble learning \cite{gong2022preserving}, and prompt learning \cite{su2024federated}. To mitigate fine-grained domain heterogeneity, an effective strategy is to combine contrastive loss to map local pixel embeddings into a global semantic space \cite{miao2023fedseg,tan2024bridging} and learn global class embeddings. Inaccurate mapping often occurs, leading to inconsistencies between local and global representations. To tackle this challenge, we introduce class exemplars to explore both intra-class and inter-class relationships, which are then used to supervise the client-side global branches, ensuring alignment with global-level representations in the semantic space.  

\section{Methodology}
\subsection{Overview}
Figure~\ref{fig2}(a) illustrates the overall framework of the proposed FedSaaS approach. The training objective is to ensure that the local and global representations of each class, derived from all client datasets, are mapped into a shared semantic space. For a total of \( N \) clients, each client holds its own dataset (\( \mathcal{D} _1, \dots, \mathcal{D} _N \)). To maintain consistency between local and global representations for each class, we train a model \( \mathcal{F}  \) on each client (\( \mathcal{F} _1, \dots, \mathcal{F} _N \)), with collaborative training occurring across them. Specifically, each model \( \mathcal{F}  \) consists of two branches: a global branch \( \mathcal{F} ^G \) and a local branch \(\mathcal{F} ^L\). Although the data used by the two branches partially overlap, their representations remain inconsistent. Our goal is to align the representations at both levels by mapping them into a unified semantic space, thereby enabling harmonized outputs.

We follow the model decoupling idea proposed by  \cite{collins2021exploiting}, dividing the backbone into two components: 1) a feature extractor \( \bf{W}: \mathbb{R}^\mathit{D} \to \mathbb{R}^\mathit{K} \), which maps input samples to the feature semantic space, and 2) a segmentation head \( \bf{\Theta}: \mathbb{R}^\mathit{K}\to \mathbb{R}^\mathit{C} \), which maps feature semantic space to label space. The final fully connected layer in a given backbone network is treated as the segmentation head. Here, the parameters \( D \), \( K \), and \( C \) represent the dimensions of the input, feature, and label spaces, respectively. 

The FedSaaS training process involves operations on both the client and server sides. On the client side, the training dataset consists of raw image data. On the server side, the training dataset comprises class exemplars uploaded from clients. These exemplars are obtained by multiplying the output of the fully convolutional network (FCN) with the mask image corresponding to each class in the original image. To ensure compatibility for this multiplication, the output size of the FCN must match the size of the original image. This process is formally defined as: \(m^c_i = \text{FCN}(x) \odot M^c_i\), where \( x \in \mathbb{R}^{H \times W} \) represents the original image, and \( M^c_i \in \mathbb{R}^{H \times W} \) denotes the mask of a specific class, and \( \odot \) denotes the Hadamard product. Class exemplars capture the spatial distribution and correlations of the corresponding class, making them a valuable criterion for achieving alignment between local- and global-level class representations.

\subsection{Weighted Ensemble Distillation}
On the server side, a global branch \( \mathcal{F}^G \) is trained using weighted ensemble distillation based on class exemplars to map and constrain the global class representation. For the global branch \( \mathcal{F}_k^G \) of client \( k \), when provided with unseen exemplars \( m_i^c, i\neq k \), the predicted logits are computed as \( z_k = \mathcal{F}_k^G({\bf{W}}_k^G, {\bf{\Theta}}_k^G; m_i^c) \). Simultaneously, the server-side global branch \( \mathcal{F}^G \) generates predictions for the same exemplars, producing logits \( z = \mathcal{F}^G({\bf{W}}^G, {\bf{\Theta}}^G; m_i^c) \). The similarity between the outputs of the client-side and server-side global branches is measured by \( \text{sim}(z_k, z) \). A higher value of \( \text{sim}(z_k, z) \) indicates that the client’s global branch exhibits better generalization to unseen domains, and thus, it is assigned a higher weight. To quantify this, we define a weight function \( \alpha_i \) that reflects the generalization capability of each client’s global branch: \(\alpha_i = \frac{\text{sim}(z_k, z)}{\sum_{i=1, i \neq k}^N \text{sim}(z_k, z)}\), where the similarity measure \( \text{sim}(\cdot) \) can be implemented using metrics such as Kullback-Leibler divergence or cosine similarity. 

Subsequently, the logits from the client-side global branches are aggregated in a weighted manner to produce new predicted logits: \( \hat{z}=\sum_{i=1}^N \alpha_i \cdot z_i \). The server-side global branch is then trained by minimizing the following loss function designed to learn from the aggregated logits: 
\begin{equation} \nonumber
    \mathcal{L}_{distill} = \mathbb{E}_{m^c} [\text{sim}(z,\hat{z})].
\end{equation}
This process enables the server-side global branch to effectively integrate knowledge from multiple clients, thereby achieving enhanced generalization across diverse domains. 

To enhance the constraint of global semantic consistency, we introduce inter-client contrastive learning on the server side, leveraging class exemplars. Class exemplars encapsulate semantic information specific to client categories, providing a direct basis for constructing positive and negative samples in multi-level contrastive learning. We define the inter-client contrastive loss as: 
\begin{align} \nonumber
\mathcal{L}_{inter} = -\frac{1}{N} \sum_{i=1}^N \log \frac{\exp\left({ v_i^c v^+ }/{\tau}\right)}{\exp\left({ v_i^c v^+ }/ {\tau}\right) + \sum_{v^-} \exp\left({ v_i^c v^- }/{\tau}\right)},
\end{align}
where \( v_i^c \) is the normalized vector of the class exemplar \( m_i^c \), \( v^+ \) represents positive samples from the same class, \( v^- \) represents negative samples from other classes, and \( \tau \) is the temperature coefficient. The overall loss function for training the server-side global branch is 
\begin{align}
\mathcal{L}_g = \mathcal{L}_{distill} + \mathcal{L}_{inter}.
\end{align}%
By training the global branch from the client to the server, the consistency of similar representations is improved, ensuring aligned semantic representations across clients. 

\subsection{Global Prototype Supervision}
The server further generates class prototypes to supervise the client-side global branch, ensuring alignment between local- and global-level class representations. As illustrated in Figure~3, these class prototypes are designed based on the deep representations of class exemplars, denoted as \(h^c(x,y)\). 

To construct the class prototypes, we generate category distribution vectors and category co-occurrence relationships separately. First, we perform a weighted average pooling across all class exemplars to obtain the category distribution vector for each class. The weight is calculated as: \(\beta_c=1-K^{-1}_c/max(K^{-1}_{c'})-min(K^{-1}_{c'})\), where \(K_c\) represents the number of class exemplars for class \(c\) and \(K^{-1}_c\) denotes its inverse value. This weighting scheme addresses the issue of underrepresentation for classes with fewer exemplars during training by assigning them larger weights. Based on the deep representations \(h^c(x,y)\) and  the computed weights, the class distribution vector \( v_c \) for each class is calculated as: 
\begin{align}
\mathbf{v}_c = \Phi\left[\frac{1}{{K_c}} \sum_{k=1}^{K_c} \beta_c \cdot h^c(x,y)]\right],
\end{align}
where \(\Phi[\cdot]\) represents a dimensionality reduction operation. 

Next, we generate the co-occurrence relationship \( \phi_{c, c'} \) between class \( c \) and class \( c' \), which reflects their relative distribution across all spatial positions. It is computed as 
\begin{align}
\phi_{c, c'} = \frac{\textstyle \sum_{ \Omega_c} \textstyle \sum_{ \Omega_{(c, c')}} \mathbb{I} [h^c(x,y) \neq 0 \wedge h^{c'}(x',y') \neq 0]}{\textstyle \sum_{ \Omega} \mathbb{I} [h^c(x,y) \neq 0]},
\end{align}
where \( \Omega_{(c, c')} \) represents the neighborhood region of classes \( c \) and \( c' \), and \( \mathbb{I}[\cdot] \) denotes the indicator function. Based on the co-occurrence relationship, we further incorporate positional information and regional relationships of class exemplars to calculate the correlation \( R_{c, c'} \) between classes \( c \) and \( c' \):
\begin{align}
R_{c, c'} = &\phi_{c, c'} \cdot \sum\limits_{\Omega_c} h^c(x,y) \cdot \nonumber\\ \cdot &\left(\sum\limits_{\Omega_{(c, c')}} (h^c(x',y')) \cdot K_d(x,y,x',y') \right),
\end{align}
where \( K_d(\cdot) \) represents a Gaussian kernel function that weights the influence of distance between adjacent pixels. 

By combining the distribution vector $\mathbf{v}_c$ and the correlation $R_{c,c'}$ for each class, we obtain the class prototype \( \mathbf{g}_c \):
\begin{align}
\mathbf{g}_c = \mathbf{v}_c + \frac{1}{|C|} \sum\limits_{c' \in C, c' \neq c} R_{c,c'} \cdot v_{c'},
\end{align}
where $|C|$ denotes the total number of classes. 

Class prototypes contain two key features: intra-class distribution and inter-class co-occurrence relationship. Once the clients receive the class prototypes \( \mathbf{g}_c \), these prototypes are converted into dynamic weights for the convolutional kernels in the global branch. To achieve this, we define a weighted generation network \( f_{\mathbf{w}}(\mathbf{g}_c) \), which projects \( \mathbf{g}_c \) into the weight dimension. This network can be implemented using a multilayer perceptron (MLP). Through the supervision of class prototypes, the local-level class representations generated by the client-side global branch are effectively aligned with the global representations, ensuring consistency across domains. 

\begin{figure}[!t]
\centering
\includegraphics [width=\columnwidth] {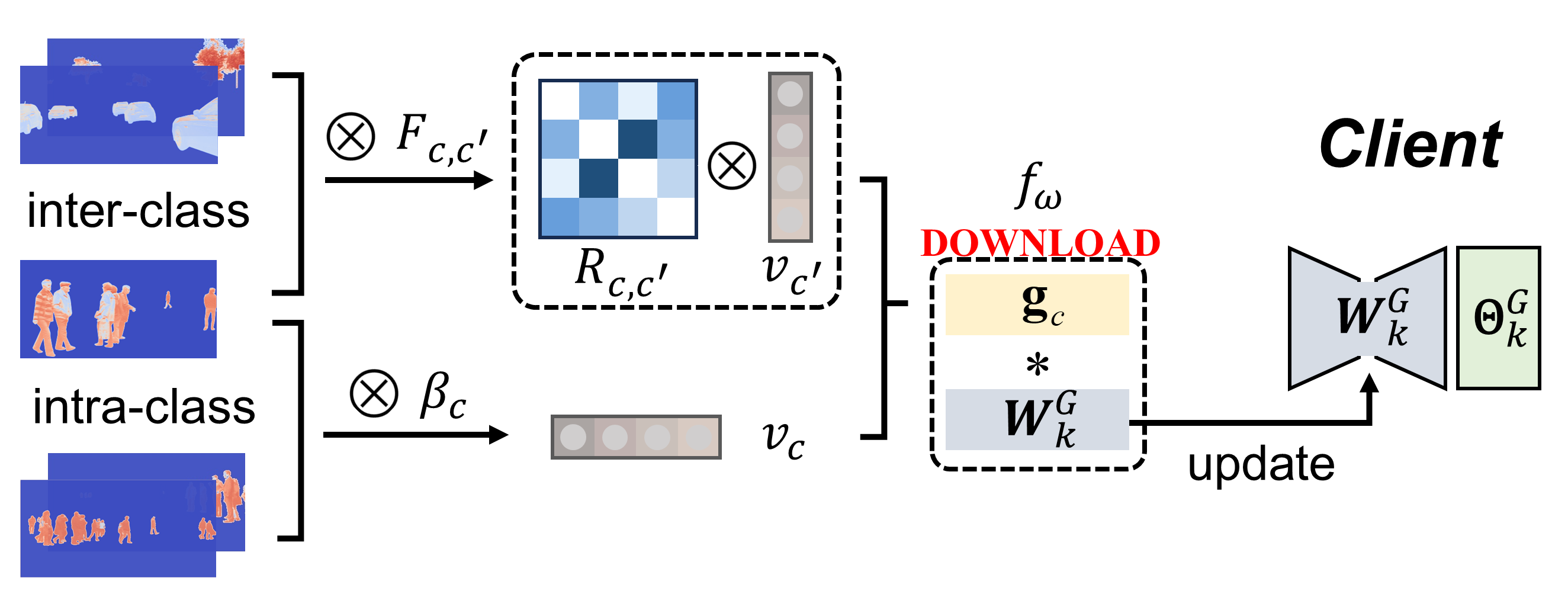}
\caption{Diagram of the prototype head and process of global prototype supervision.}
\label{fig3}
\end{figure}

\subsection{Local Adversarial Harmonization}
On the client side, the local branch and global branch focus on different levels of class representations, necessitating a mechanism to balance their contributions to the final model output. To address this, we propose a local adversarial harmonization mechanism that facilitates mutual learning between the two branches by confusing a newly trained discriminator. This mechanism is divided into two stages: discriminator training and branches training. 

\textbf{Discriminator training.} Given the output logits of the local and global branches, a domain discriminator is trained to distinguish their origins. The objective is to maximize the discriminator's classification accuracy, ensuring it can correctly differentiate between the outputs of the local and global branches. To this end, the discriminator maximizes a binary cross-entropy loss function: 
\begin{align}
\mathcal{L}_d = -\mathbb{E}_{d} [p\log \hat{p} + (1-p)\log (1 - \hat{p})],
\end{align}
where \( p \in \{0, 1\} \) represents the ground-truth label, indicating whether the current input originates from the local branch (\( p = 0 \)) or the global branch (\( p = 1 \)), and \( \hat{p} \) is the prediction output of the domain discriminator.

\textbf{Branches training.} To counter the discriminator, the training objective of the local and global branches is to make the domain discriminator incapable of distinguishing their outputs. This is achieved by incorporating adversarial constraints into their loss functions, which aim to minimize the discriminator loss and thereby deceive the discriminator. For the global branch, we adjust the parameters of the feature extractor under the supervision of global prototypes. The parameters of the segmentation head are first replaced with the server-side parameters ${\bf{\Theta}}^G$, and subsequently optimized using a segmentation loss \(\mathcal{L}^{g}_{seg} \). 
The training losses for the two branches are defined as: 
\begin{equation}
\begin{aligned}
&\mathcal{L}_{global} = \mathcal{L}^{g}_{seg} + \lambda \mathcal{L}_d,\\
&\mathcal{L}_{local} = \mathcal{L}^{l}_{seg} + \lambda \mathcal{L}_d + \mathcal{L}_{intra},
\end{aligned}
\end{equation}
where \( \mathcal{L}^{l}_{seg} \) is the local segmentation loss and $\lambda$ is a hyperparameter that adjusts the weight of discriminator loss $\mathcal{L}_d$. A contrastive loss, $\mathcal{L}_{intra}$, derived from the client's own class exemplars, is added to the local branch loss to constrain the local-level class representations: 
\begin{align}
\mathcal{L}_{intra} = -\log \frac{\exp ( v^c v^+/{\tau})}{\exp (v^cv^+/{\tau}) + \sum_{v^-}\exp({ v^c v^- }/{\tau})}.
\end{align}

The logits from the two branches are summed and averaged to produce the final output. 
Through the local adversarial mechanism, the segmentation model dynamically harmonizes the contributions of local and global representations, ensuring consistent outputs.

\section{Experiments}
\subsection{Experimental Settings}
To evaluate the performance of the proposed method, we select the driving scene segmentation task for both training and validation. This task involves a wide range of complex categorical objects and presents real-world class imbalance issues, posing significant challenges in achieving semantic consistency across categories. We construct datasets for two heterogeneity scenarios: slight and severe, based on training difficulty. These datasets differ notably in terms of domain shift and label shift. For the slight heterogeneity scenario, we use the widely adopted baseline dataset, Cityscapes \cite{cordts2016cityscapes}, which includes street views from multiple cities across Europe. Due to small geographical and device-related differences, this dataset exhibits minimal domain shift. For the severe heterogeneity case, we utilize five driving scene datasets: Cityscapes, Mapillary Vistas \cite{neuhold2017mapillary}, BDD100K \cite{yu2020bdd100k}, GTA5 \cite{richter2016playing}, and Synthia \cite{ros2016synthia}. This collection not only contains street views from cities worldwide but also includes simulated datasets captured from various angles and devices, resulting in both severe domain and label shifts. In this scenario, the model's ability to generalize across diverse scenes is more rigorously tested. 

\begin{figure*}[!t]
\centering
\includegraphics [width=\textwidth] {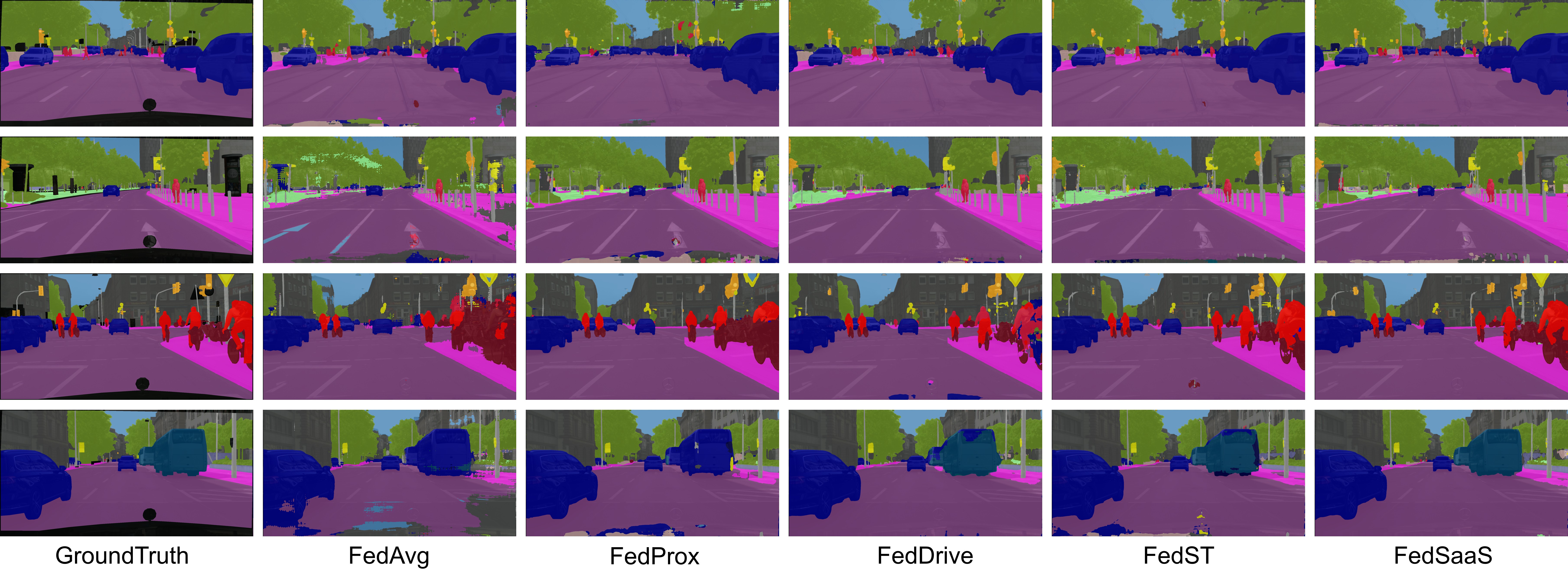}
\caption{Comparison of visual results from different methods on datasets with severe heterogeneity.}
\label{fig4}
\end{figure*}

\begin{table*}[ht]
\centering
\begin{tabular}{lccccccc}
\toprule
\multirow{2}{*}{\centering \multirow{2}*{Method}} & \multicolumn{2}{c}{Slight Heterogeneity} & \multicolumn{2}{c}{Severe Heterogeneity} & \multicolumn{2}{c}{Unseen-Domain} \\
\cmidrule(lr){2-3} \cmidrule(lr){4-5} \cmidrule(lr){6-7}
& Acc $\pm$ std & mIoU $\pm$ std & Acc $\pm$ std & mIoU $\pm$ std & Acc $\pm$ std & mIoU $\pm$ std \\
\midrule
FedAvg (PMLR 2017) & 79.20 $\pm$ 1.96 & 47.92 $\pm$ 1.53 & 66.58 $\pm$ 2.58 & 37.19 $\pm$ 2.12 & 61.61 $\pm$ 1.68 & 34.28 $\pm$ 2.04 \\
FedProx (MLSys 2020) & 78.93 $\pm$ 0.67 & 47.62 $\pm$ 0.90 & 67.06 $\pm$ 1.26 & 38.08 $\pm$ 1.62 & 63.09 $\pm$ 1.91 & 35.37 $\pm$ 1.54 \\
FedDrive (IROS 2022) & 84.74 $\pm$ 1.31 & 51.14 $\pm$ 0.77 & 73.24 $\pm$ 1.31 & 42.83 $\pm$ 0.56 & 70.40 $\pm$ 1.85 & 43.06 $\pm$ 0.78 \\
FedSeg (CVPR 2023) & 84.36 $\pm$ 1.81 & 51.36 $\pm$ 1.29 & 72.86 $\pm$ 2.89 & 40.14 $\pm$ 2.04 & 67.04 $\pm$ 2.25 & 37.48 $\pm$ 2.30 \\
FedST (AAAI 2024) & 85.32 $\pm$ 0.68 & 52.60 $\pm$ 0.36 & 75.58 $\pm$ 1.99 & 44.19 $\pm$ 1.37 & 73.43 $\pm$ 1.89 & 44.77 $\pm$ 0.96 \\
FedSaaS (Ours) & \textbf{88.57 $\pm$ 0.95} & \textbf{54.34 $\pm$ 0.62} & \textbf{82.26 $\pm$ 2.30} & \textbf{48.67 $\pm$ 1.18} & \textbf{74.15 $\pm$ 1.47} & \textbf{45.96 $\pm$ 1.14} \\
\midrule
Backbone & 81.09 $\pm$ 0.73 & 49.61 $\pm$ 0.58 & 68.14 $\pm$ 1.77 & 37.55 $\pm$ 0.93 & 65.24 $\pm$ 1.66 & 35.63 $\pm$ 1.47 \\
+ \textit{Proto.} & 85.53 $\pm$ 1.04 & 53.57 $\pm$ 0.73 & 76.06 $\pm$ 2.49 & 43.08 $\pm$ 1.72 & 70.08 $\pm$ 1.36 & 39.08 $\pm$ 1.25 \\
+ \textit{Proto.} + \(\mathcal{L}_{multi-con}\) & 88.21 $\pm$ 1.27 & 54.21 $\pm$ 0.66 & 80.38 $\pm$ 2.61 & 46.21 $\pm$ 1.46 & 73.96 $\pm$ 1.86 & 45.83 $\pm$ 1.30 \\
+ \textit{Proto.} + \(\mathcal{L}_{multi-con}\) + \(\mathcal{L}_d\) & \textbf{88.57 $\pm$ 0.95} & \textbf{54.34 $\pm$ 0.62} & \textbf{82.26 $\pm$ 2.30} & \textbf{48.67 $\pm$ 1.18} & \textbf{74.15 $\pm$ 1.47} & \textbf{46.06 $\pm$ 1.14} \\
\bottomrule
\end{tabular}
\caption{Performance comparison (\%) under slight and severe heterogeneity scenarios across various methods (top), and effectiveness validation of FedSaaS modules (bottom).} \label{per_com}
\end{table*}

We select representative FL methods and state-of-the-art (SOTA) federated segmentation approaches for comparison, including FedAvg \cite{mcmahan2017communication}, FedProx \cite{li2020federated}, FedDrive \cite{fantauzzo2022feddrive}, FedSeg \cite{miao2023fedseg}, and FedST \cite{ma2024fedst}. Due to differences in dataset size and annotation types, we standardize the data preprocessing by cropping all images and resizing them to \(512 \times 1024\). For model training, we adopt the BiSeNet V2 architecture \cite{yu2021bisenet}, a lightweight network designed to capture both spatial features and high-level semantic context. The temperature coefficient \( \tau \) for the multilevel contrastive loss and the weight \( \lambda \) for the adversarial loss are set to $0.05$ and $0.1$, respectively. We use a batch size of $16$, with $10$ local iterations and $50$ communication rounds. 
To evaluate performance, we use two common semantic segmentation metrics: mean Intersection over Union (mIoU), which measures the intersection over union between predicted and ground truth pixels averaged across all categories, and Pixel Accuracy, which describes the ratio of correctly classified pixels. 

\subsection{Main Results}
\textbf{Quantitative Analysis.} We evaluate the performance on validation datasets, reporting the average and fluctuation deviation over three independent tests. Table~\ref{per_com} presents a comparison between our FedSaaS method and other SOTA methods.  As shown, our method achieves the best performance across all data environments. In the slight heterogeneity scenario, we observe a modest improvement over the current best methods, with increases of 2.75\% in Accuracy and 1.66\% in mIoU. In the severe heterogeneity case, FedSaaS performs significantly better, with improvements of 6.68\% in Accuracy and 4.48\% in mIoU. This is because FedSaaS leverages class consistency to jointly represent the characteristics of classes at both the local and global levels. Additionally, to assess generalization, we separate Cityscapes from the severely heterogeneous dataset and use it as an unseen domain for testing. We average the results of the remaining four clients, and the scores demonstrate that FedSaaS outperforms the SOTA methods in this scenario as well. 

\textbf{Qualitative Analysis.} Figure~4 presents the visualization results of testing different clients under the severe heterogeneity scenario. It is observed that Feddrive and FedST, both designed to address the domain shift problem, improve segmentation accuracy only for certain classes compared to traditional FL methods. In contrast, our FedSaaS method improves segmentation performance across all categories. By aligning and constraining both local and global class representations, our approach ensures high segmentation performance for all clients. More detailed comparisons are shown in the Appendix of full version of this paper.

\textbf{Ablation Studies.} We evaluate the effectiveness of each module, as shown in the lower half of Table~\ref{per_com}. The initial backbone structure consists of local and global branches on client side and follows the FedAvg configuration. The outputs from both branches are combined through summation and averaging to produce the final output. Subsequently, we incorporate class exemplars into the backbone, enabling the segmentation head ${\bf{\Theta}}^G$ to be trained on server side while simultaneously generating class prototypes (abbreviated as \textit{Proto.} in the subsequent charts) to supervise parameter updates in client-side global branch. This enhancement significantly improves the performance of the overall framework, resulting in an accuracy increase ranging from 4.42\% to 7.92\%. 

We further incorporate a multilevel contrastive loss (abbreviated as \(\mathcal{L}_\mathrm{multi-con}\) in the table) based on class exemplars to enforce class representation constraints at both the local and global levels. Compared to the results from the slight heterogeneity scenario, the combination of prototype supervision and multilevel contrastive loss yields a particularly notable performance improvement in the severe heterogeneity case, with an increase of 12.24\% in Accuracy and 8.66\% in mIoU. This highlights the importance of achieving class consistency in scenarios with significant domain shifts. We also introduce an adversarial harmonization module before the outputs of the two client branches. The purpose of this module is to balance the contributions of local and global semantic representations, leading to more consistent model outputs. By comparing the results before and after the inclusion of this module, we observe an enhancement in the model’s performance. Notably, the alignment between global and local representations is achieved through the above mechanisms, which enables the adversarial mechanism to operate effectively on both branches.

\subsection{Empirical Analysis}
\begin{figure}[h]
\centering
\includegraphics [width=\columnwidth] {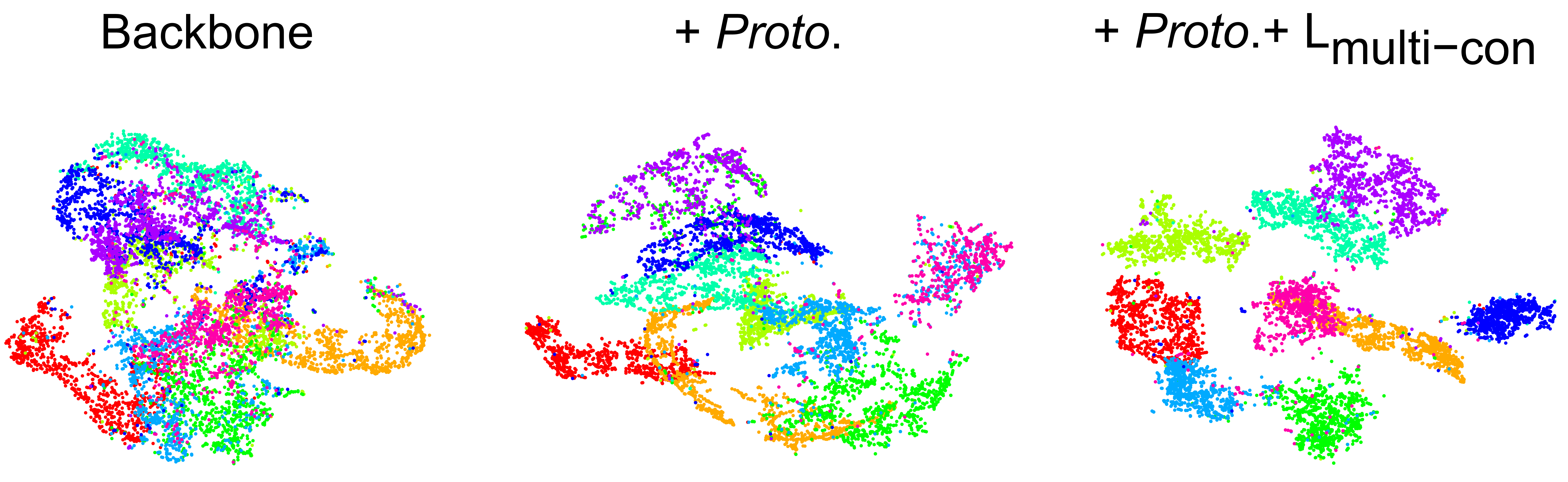}
\caption{Visualization of the pixel embeddings for differenct semantic classes.}
\label{fig5}
\end{figure}

\textbf{Visualization.} We use t-SNE \cite{van2008visualizing} to visualize the pixel embeddings of semantic classes under the severe heterogeneity scenario, as shown in Figure~\ref{fig5}. 
The results demonstrate that, without additional modules, the model produces poor embeddings, with most semantic class pixels intermingled. Incorporating the \textit{Proto.} module results in a noticeable separation between pixels of different semantic classes. The introduction of the two-level contrastive loss (\(\mathcal{L}_{multi-con}\)) further enhances the divergence between categories in the embedding space, highlighting its critical role in constraining semantic class representations. 

\begin{figure}[h]
\centering
\includegraphics [width=\columnwidth] {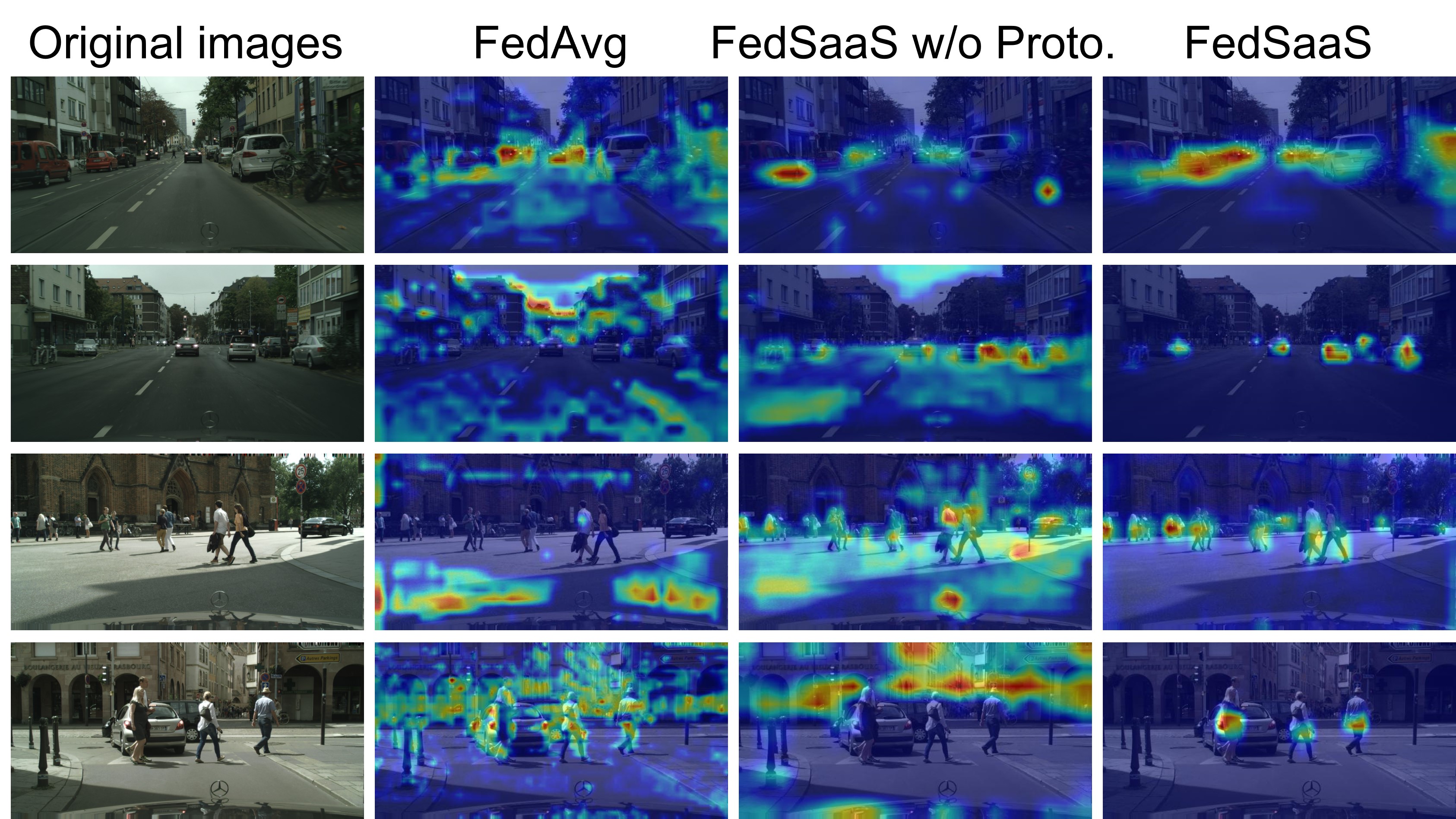}
\caption{Grad-CAM visualization of the feature attention regions in the last convolutional layer of the global branch.}
\label{fig6}
\end{figure}

We visualize the global branch's attention to categories using Grad-CAM \cite{selvaraju2017grad}. As shown in Figure~\ref{fig6}, we compare the attention given by the model to the most common categories in driving scenes---namely, pedestrians and vehicles---by analyzing the output of the last convolutional layer in both the FedAvg and FedSaaS global branches. It is evident that both FedAvg and FedSaaS w/o \textit{Proto.} exhibit insufficient focus and attention to the image. After incorporating the \textit{Proto.} module, the global branch is able to accurately locate and identify the corresponding categories in the respective channels. This supervision of local class alignment through global class prototypes enables the model to gain a richer understanding of various categories, which is crucial for achieving high segmentation accuracy. 

\begin{figure}[h]
\centering
\includegraphics [width=\columnwidth] {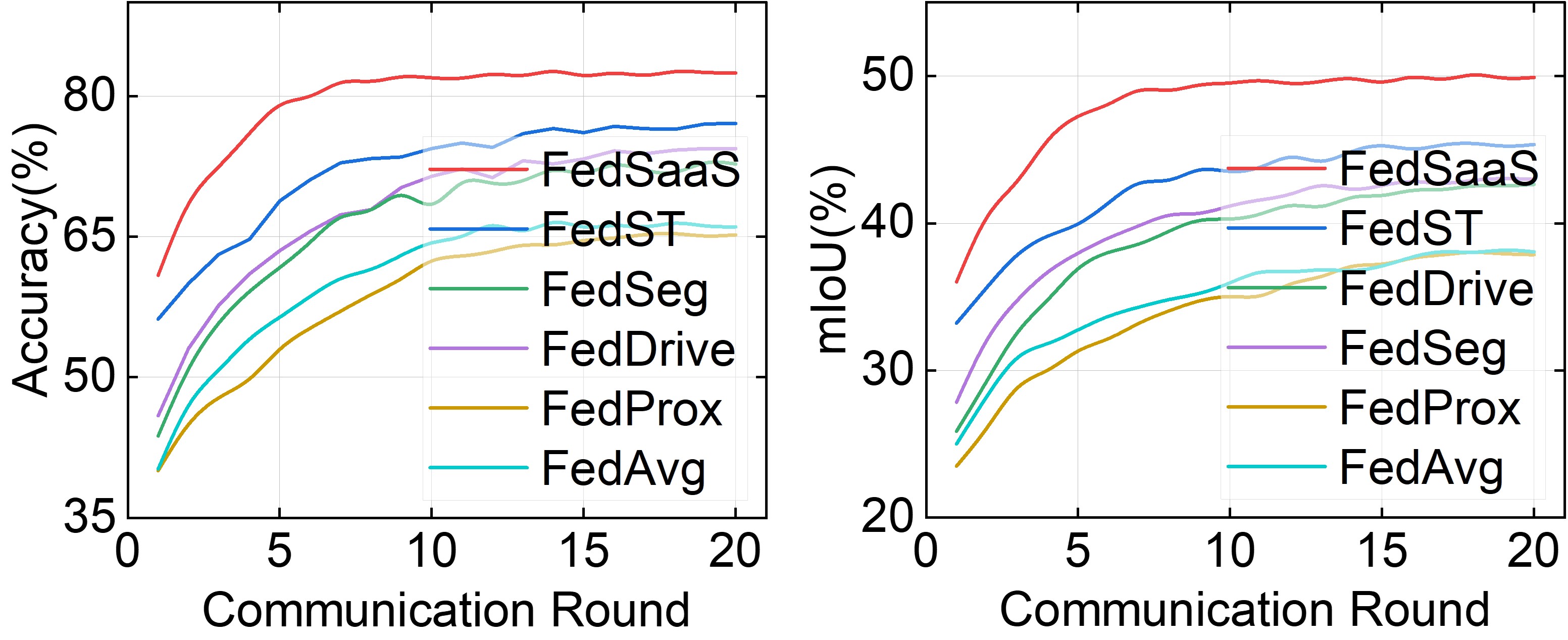}
\caption{Comparison of communication efficiency.}
\label{fig7}
\end{figure}

\begin{table}[ht]
\centering
\begin{tabular}{c|cc|cc}
\toprule
\multicolumn{1}{c|}{Upload} & \multicolumn{2}{c}{Slight Heterogeneity} & \multicolumn{2}{|c}{Severe Heterogeneity} \\
ratio& Acc   & mIoU  & Acc   & mIoU  \\
\midrule
25\% & 84.26 & 51.27 & 69.12 & 35.97 \\
50\% & 86.53 & 53.84 & 75.50 & 43.28 \\
75\% & 87.96 & 54.08 & 77.74 & 44.36 \\
\bottomrule
\end{tabular}
\caption{Performance comparison (\%) of different upload ratios of class exemplars under slight and severe heterogeneity scenarios.} \label{table_2}
\end{table}

\textbf{Communication Efficiency.} Figure~\ref{fig7} illustrates the performance curves of different methods during training. It is observed that, compared to other methods within the same communication round, FedSaaS achieves superior performance with enhanced communication efficiency. Additionally, the need to upload class exemplars from clients introduces extra communication overhead. We examine the impact of randomly uploading a subset of class exemplars on performance. As shown in Table~\ref{table_2}, as the number of uploaded exemplars increases, segmentation accuracy improves. Notably, test accuracy when uploading approximately 50\%-75\% of the exemplars is significantly higher than when uploading only 25\%. This suggests that, in scenarios where the dataset is large or communication constraints are tight, uploading around half of the class exemplars can still achieve relatively high accuracy, thereby reducing transmission costs.

\textbf{Stability Analysis}.
The branch and discriminator are optimized toward opposing objectives: the branch aims to minimize its loss, while the adversarial dynamics force the discriminator’s loss to increase. This competition may introduce initial instability in the training process, particularly under severe data heterogeneity, manifesting as transient loss fluctuations or occasional branch misalignment. To mitigate these effects, we adopt two strategies: 1) We initialize the adversarial weight $\lambda$ at a reduced value to limit early-stage perturbation. As optimization stabilizes, $\lambda$ is progressively increased to its target value. 2) During training, sharp performance declines trigger a rollback to the best checkpoint, accompanied by gradient clipping on $\mathcal{L_\textbf{d}}$ to suppress instability. As shown in Table~\ref{per_com} and Figure~\ref{fig7}, these measures maintain stable performance growth without compromising convergence speed.

\section{Conclusion}
In this paper, we have proposed a class-consistency FL approach tailored for semantic segmentation tasks. To address the ambiguity in class representations caused by domain shifts, we have introduced a novel framework, FedSaaS, which leverages class exemplars as a criterion to ensure consistency between local and global class representations. Specifically, we have sequentially incorporated class prototypes on the server side and an adversarial mechanism on the client side to achieve two-level representation alignment, thereby ensuring consistent outputs across clients. Extensive experiments conducted on five driving scene datasets have demonstrated that FedSaaS outperforms state-of-the-art methods in addressing the class-consistency representation problem. In future work, we aim to investigate the federated semantic segmentation problem further, taking into account both domain shift and model heterogeneity factors. 

\clearpage

\appendix
\subfile{appendix.tex}

\clearpage
\bibliographystyle{named}
\bibliography{ijcai25}
\end{document}

%% file: appendix.tex
\section*{Appendix}

\section{Algorithm Details}
In this section, we supplement the pseudocode of FedSaaS we missed in the main text, which is given in Algortihm~\ref{alg:algorithm}.
\begin{algorithm}
    \caption{FedSaaS}
    \label{alg:algorithm}
    \textbf{Input:} $N$ clients with local datasets (\( \mathcal{D} _1, \dots, \mathcal{D} _N \)), global model parameters $ \mathcal{F}^G ( \mathcal{\mathbf{\Theta}}^{G}\, \mathcal{\mathbf{W}}^{G})$, local model parameters $ \mathcal{F}^L (\mathcal{\mathbf{\Theta}}^{L}, \mathcal{\mathbf{W}}^{L}) $, total communication rounds $T$, temperature coefficient $ \tau$, and adversarial loss weight $ \lambda$.\\
    \textbf{Output:} $ \{\mathcal{F}_1, \dots, \mathcal{F}_N \}$.
    \begin{algorithmic}[1]
        \STATE Initialize parameters: global and local branch $ \mathcal{F}^{G,0} $ and $ \mathcal{F}^{L,0} $ for each client.
        \STATE Generate class exemplars $m_i^c$  and upload to the server.
        \FOR{each communication round $t \in \{1, \dots, T\}$}
            \STATE Sever send class prototype $g_{c}^t$ and segmentation head $ \mathbf{\Theta}_i^{G,t}$ to the selected clients.
            \FOR{client $i$ in parallel}
                \STATE Overwrite client-side segmentation head $ \mathbf{\Theta}_i^{G,t} $ with the server-side segmentation head $ \mathbf{\Theta}^{G,t}$.
                \STATE Supervise client-side extractor $\mathbf{W}_i^{G,t}$ with weighted generation function $f(g_c^t)$.
                \STATE Update discriminator by Eq. (6).
                \STATE  Update  client-side global and local branch $\mathcal{F}_i^t(\mathcal{F}_i^{L,t}, \mathcal{F}_i^{G,t})$ by Eq. (7).
            \ENDFOR
            \STATE Upload client-side global branch $ \mathcal{F}_i^{G,t}$ to the server.
        \ENDFOR
        \STATE Aggregate and calculate server-side global branch: $ \mathcal{F}^G = \frac{1}{n}\sum_{i \in C_t} \mathcal{F}_i^G$.
        \STATE Update server-side global branch $\mathcal{F}^G$.
        \STATE Generate class prototypes $g_c$ by Eq.~(2)--Eq.~(5).
        \STATE \textbf{Return:} Client segmentation model $ \{ \mathcal{F}_1, \dots, \mathcal{F}_N \}$.
    \end{algorithmic}
\end{algorithm}

\section{Dataset and Missing Implementation Details}
To evaluate the performance of cross-domain data across different methods, we construct a highly heterogeneous dataset comprising both real-world driving scene data and virtual synthetic data. For the experiments, we utilize five driving scene datasets for evaluation. The details of these datasets are as follows:

\begin{figure}[h]
\centering
\includegraphics [width=\columnwidth] {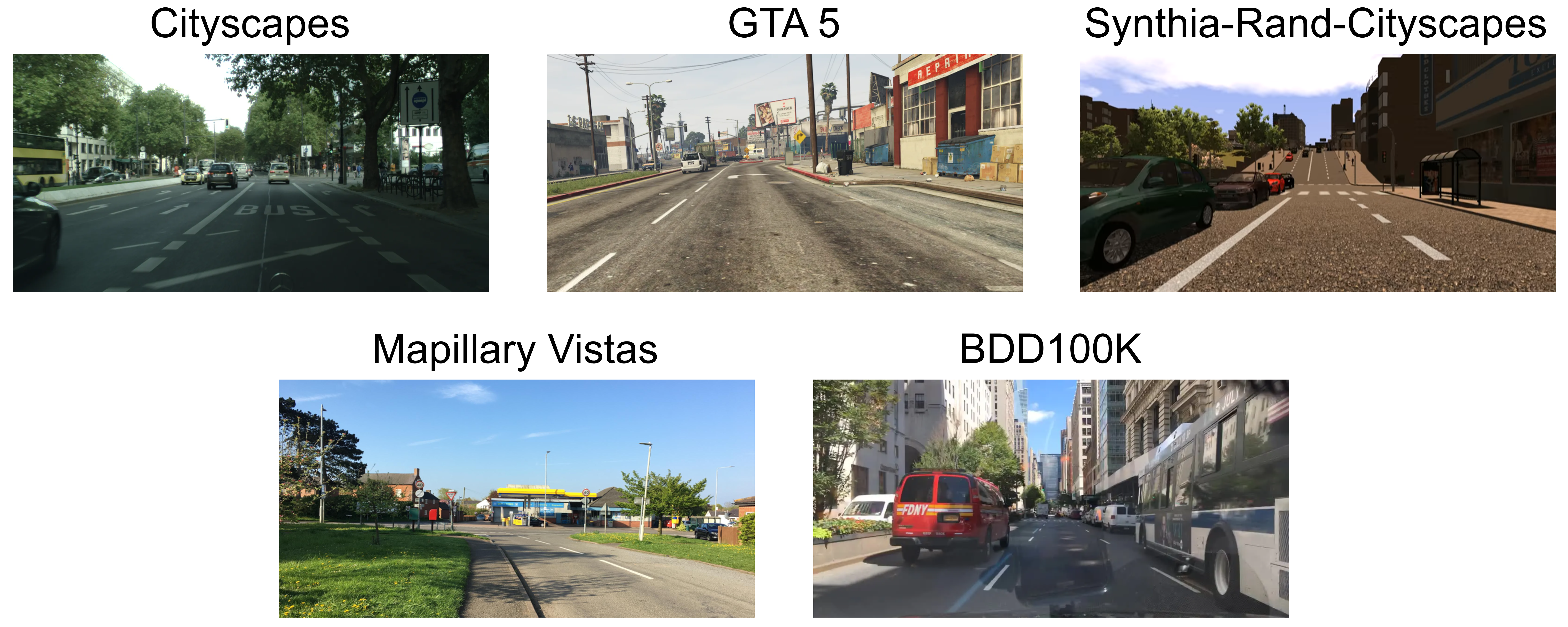}
\caption{The highly heterogeneous dataset consists of five datasets, each exhibiting distinct appearances.}
\label{fig1}
\end{figure}

\begin{figure}[h]
\centering
\includegraphics [width=\columnwidth] {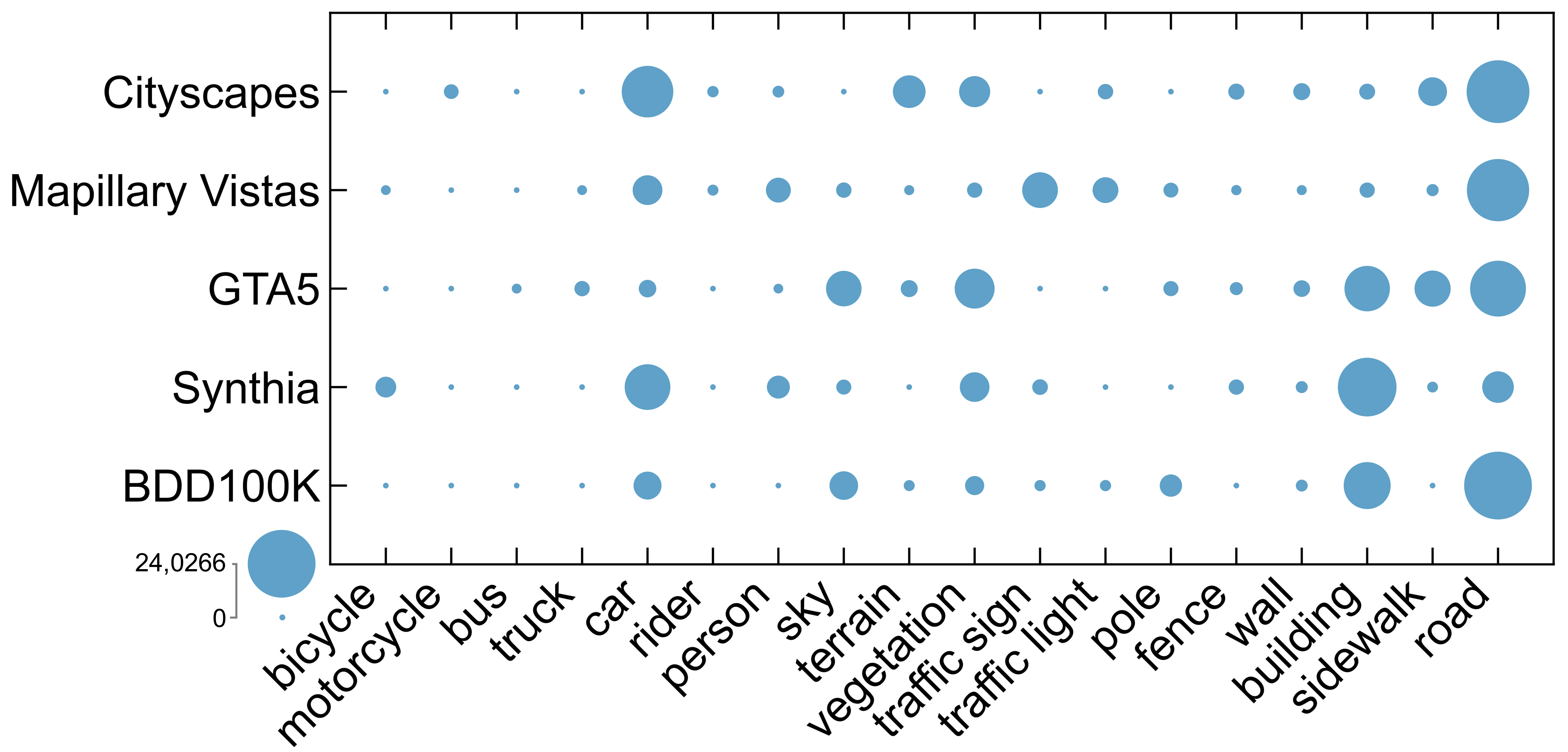}
\caption{Label distribution bubble chart for the five datasets.}
\label{fig2}
\end{figure}

\textbf{Cityscapes} \cite{cordts2016cityscapes}: A large-scale dataset designed for urban scene analysis and computer vision tasks. Created by the Computer Vision Lab at Darmstadt University of Technology, the dataset comprises images collected from 50 different cities across Germany, including 5,000 high-resolution images with detailed pixel-level semantic annotations. It is well-suited for tasks such as semantic segmentation and instance segmentation. 

\begin{table*}[ht!]
\centering
\begin{tabular}{c|l|ccccccc}
\hline
\multirow{2}{*}{Dataset} & \multirow{2}{*}{Category} & \multicolumn{7}{c}{Methods} \\
\cline{3-9}
 & & Centralized & FedAvg & FedProx & FedDrive & FedSeg & FedST & FedSaaS \\
\hline
\multirow{7}{*}{Cityscapes} & person & 70.41 & 56.64 & 59.08 & 62.39 & 61.95 & 61.65 & \textbf{66.41} \\
 & car & 86.80 & 77.29 & 79.54 & 83.28 & 80.33 & 82.93 & \textbf{88.84} \\
 & bicycle & 65.44 & 55.36 & 53.47 & 57.62 & 58.44 & \textbf{62.93} & 61.99 \\
 & traffic sign & 64.94 & 51.61 & 51.82 & 56.62 & 55.67 & 58.33 & \textbf{60.53} \\
 & vegetation & 38.11 & 26.93 & 28.56 & 32.85 & 30.01 & 33.07 & \textbf{36.66} \\
 & road & 80.34 & 70.63 & 69.67 & \textbf{76.77} & 71.94 & 75.93 & 74.51 \\
 & Mean & 67.67 & 56.41 & 57.02 & 61.59 & 60.17 & 62.33 & \textbf{64.16} \\
\hline
\multirow{7}{*}{Mapillary Vistas} & person & 64.42 & 54.51 & 48.97 & 55.89 & 52.06 & 56.44 & \textbf{58.35} \\
 & car & 73.93 & 57.57 & 59.42 & 64.91 & 60.51 & 64.77 & \textbf{65.04} \\
 & bicycle & 57.34 & 42.09 & 40.32 & 51.48 & 46.79 & 49.56 & \textbf{52.74} \\
 & traffic sign & 49.69 & 26.34 & 29.09 & 35.57 & 25.59 & 30.73 & \textbf{40.02} \\
 & vegetation & 68.71 & 52.95 & 55.06 & 59.37 & 57.51 & \textbf{64.71} & 62.95 \\
 & road & 75.51 & 64.16 & 63.37 & 67.08 & 66.95 & 66.31 & \textbf{69.16} \\
 & Mean & 64.93 & 49.70 & 49.35 & 55.60 & 51.57 & 55.42 & \textbf{58.04} \\
\hline
\multirow{7}{*}{GTA5} & person & 64.21 & 50.37 & 52.18 & 55.77 & 59.30 & 57.04 & \textbf{60.25} \\
 & car & 77.34 & 64.81 & 62.79 & 68.29 & 64.56 & 71.82 & \textbf{76.14} \\
 & bicycle & 48.98 & 37.23 & 35.17 & 41.68 & 42.28 & 42.71 & \textbf{45.42} \\
 & traffic sign & 53.33 & 42.57 & 40.60 & 44.75 & 46.89 & 47.20 & \textbf{48.23} \\
 & vegetation & 57.48 & 40.09 & 43.85 & \textbf{53.11} & 40.76 & 46.24 & 51.31 \\
 & road & 68.26 & 57.98 & 60.15 & 65.27 & 60.86 & 62.69 & \textbf{65.98} \\
 & Mean & 61.60 & 48.84 & 49.12 & 54.81 & 52.44 & 54.67 & \textbf{57.89} \\
\hline
\multirow{7}{*}{Synthia} & person & 57.43 & 46.76 & 46.19 & 48.98 & 49.27 & 51.19 & \textbf{54.45} \\
 & car & 68.18 & 57.54 & 53.21 & 58.87 & 57.55 & 61.77 & \textbf{67.28} \\
 & bicycle & 31.89 & 21.64 & 19.57 & 26.68 & 23.79 & 26.90 & \textbf{29.07} \\
 & traffic sign & 39.78 & 28.12 & 26.65 & 34.34 & 30.23 & 35.45 & \textbf{36.67} \\
 & vegetation & 48.67 & 23.63 & 30.01 & 37.45 & 36.87 & 40.32 & \textbf{43.56} \\
 & road & 63.54 & 47.12 & 50.89 & 60.74 & 53.07 & 58.92 & \textbf{61.30} \\
 & Mean & 51.57 & 37.87 & 37.75 & 44.51 & 41.80 & 45.76 & \textbf{48.72} \\
\hline
\multirow{7}{*}{BDD100K} & person & 64.30 & 51.86 & 52.02 & 56.51 & 56.17 & 58.73 & \textbf{59.96} \\
 & car & 75.60 & 66.26 & 65.63 & 70.82 & 71.36 & 70.54 & \textbf{73.22} \\
 & bicycle & 26.37 & 11.83 & 8.44 & 14.90 & 13.49 & 16.14 & \textbf{21.64} \\
 & traffic sign & 46.08 & 33.67 & 31.29 & 39.41 & 36.58 & \textbf{42.12} & 40.71 \\
 & vegetation & 55.24 & 43.78 & 42.99 & 46.52 & 47.19 & 50.33 & \textbf{52.74} \\
 & road & 72.47 & 62.89 & 62.16 & 63.55 & 60.92 & 64.03 & \textbf{68.61} \\
 & Mean & 56.68 & 45.05 & 43.72 & 48.62 & 47.62 & 50.32 & \textbf{52.81} \\
\hline
\end{tabular}
\caption{Comparison of mIoU scores (\%) for specific classes across five clients under severe heterogeneous dataset.} \label{table_1}
\end{table*}

\textbf{Mapillary Vistas} \cite{neuhold2017mapillary}: A large-scale dataset designed for understanding urban and road scene, specifically for tasks such as semantic segmentation, instance segmentation, and object detection. Created by Mapillary, the dataset includes over 250,000 high-quality street-level images from various countries and cities around the world, with rich pixel-level annotations.

\textbf{GTA5} \cite{richter2016playing}: A large-scale dataset designed for understanding urban and road scenes, specifically for tasks such as semantic segmentation, instance segmentation, and object detection. Created by Mapillary, the dataset contains over 250,000 high-quality street-level images from diverse countries and cities worldwide, accompanied by detailed pixel-level annotations. 

\textbf{Synthia} \cite{ros2016synthia}: A synthetic dataset developed by the University of Barcelona, designed for training and evaluating autonomous driving tasks. It contains various urban road and traffic scenes, including approximately 22,000 images with detailed pixel-level semantic annotations. The dataset is widely used for tasks such as image segmentation and object detection. 

\textbf{BDD100K} \cite{yu2020bdd100k}: A large-scale driving dataset designed to advance autonomous driving technology, encompassing a wide range of driving scenes, including urban, highway, and suburban environments. Developed by the Berkeley DeepDrive team at the University of California, Berkeley, the dataset contains over 100,000 images with annotations for image-level labels, semantic segmentation, and object detection. 

As shown in Figures~\ref{fig1} and~\ref{fig2}, these datasets exhibit significant appearance and statistical heterogeneity, which pose a greater challenge for domain generalization. Our goal is to evaluate the segmentation performance across each institution as well as on the unseen images within this setup. 

\textbf{Missing Implementation Details}. The framework is implemented and optimized using PyTorch \cite{paszke2019pytorch} and Stochastic Gradient Descent (SGD). The initial learning rate and weight decay are set to $0.05$ and $5\times 10^{-4}$, respectively. All comparative experiments adhere to the same training protocol and are conducted on NVIDIA A100 GPU to ensure consistency and fairness across evaluations. 

\section{Missing Experimental Results}
To further demonstrate the performance improvement achieved through class consistency, we use a strongly heterogeneous dataset and evaluate the test results for each client. As shown in Table~\ref{table_1}, we select six out of twelve classes for mIoU comparison with FedAvg \cite{mcmahan2017communication}, FedProx \cite{li2020federated}, FedDrive \cite{fantauzzo2022feddrive}, FedSeg \cite{miao2023fedseg} and FedST \cite{ma2024fedst}. These classes include common categories such as person and car, as well as rarer ones like bicycle and traffic sign, providing a comprehensive representation of the class distribution. Notably, while our method shows slight performance fluctuations in a few categories compared to others, it significantly outperforms existing methods in most classes and in overall average performance. Furthermore, we compare the results with those from centralized learning methods. Among all the approaches, FedSaaS achieves the closest average performance to that of a centralized learning environment. 

\begin{table}[ht]
\centering
\begin{tabular}{c|c|c}
\toprule
{Method} & {Upload} & {Download} \\
\midrule
BiseNet v2  & (8.79 + 6.68)G & 1.40G \\
Deeplab v2  & (8.79 + 10.25)G & 3.18G \\
\bottomrule
\end{tabular}
\caption{Communication cost of FedSaaS under BiseNet v2 and DeepLab v2.} \label{table_2}
\end{table}

\textbf{Details of the Communication Overhead}.
We conduct a comprehensive evaluation of the communication overhead of FedSaaS. Table~\ref{table_2} presents the associated costs under two baseline methods, BiseNet v2 \cite{yu2021bisenet} and DeepLab v2 \cite{chen2017deeplab} , using a strongly heterogeneous dataset for testing. Since class exemplars from the clients need to be uploaded only once (8.79 GB), their size is of the same order of magnitude as the model parameters uploaded during training (6.68 GB and 10.25 GB, respectively). The actual overhead varies depending on the size of the dataset or baseline method used for client training. Regarding the download phase, since we only need to broadcast the class prototype and the server-side segmentation head parameters, the overhead remains minimal, which is a significant advantage over other comparison methods.

\begin{table}[ht]
\centering
\begin{tabular}{l|cc|cc}
\toprule
\multirow{2}{*}{Method} & \multicolumn{2}{c|}{Slight Heterogeneity} & \multicolumn{2}{c}{Severe Heterogeneity} \\
& Acc  & mIoU  & Acc & mIoU  \\
\midrule
w/ \( {\bf{\Theta}}^G\) & 87.42 & 54.03 & 80.84 & 47.81 \\
w/o \( {\bf{\Theta}}^G\) & 88.57 & 54.34 & 82.26 & 48.67 \\
\bottomrule
\end{tabular}
\caption{Impact of the server-side segmentation head on the performance (\%) of the client model.} \label{table_3}
\end{table}

\textbf{Effect of the Sever-side Segmentation Head \( {\bf{\Theta}}^G\)}.
We evaluate the impact of the server-side segmentation head on the performance of the client model. Experiments are conducted on both light and severe heterogeneous datasets. As shown in Table~\ref{table_3}, incorporating the projection head results in improvements of approximately +1.29\% in Accuracy and +0.59\% in mIoU. This improvement can be attributed to the fact that the segmentation head integrates global semantic information, which facilitates the model’s ability to effectively learn both local and global representations when combined with adversarial harmonization learning. 

\begin{figure}[h]
\centering
\includegraphics [width=3in] {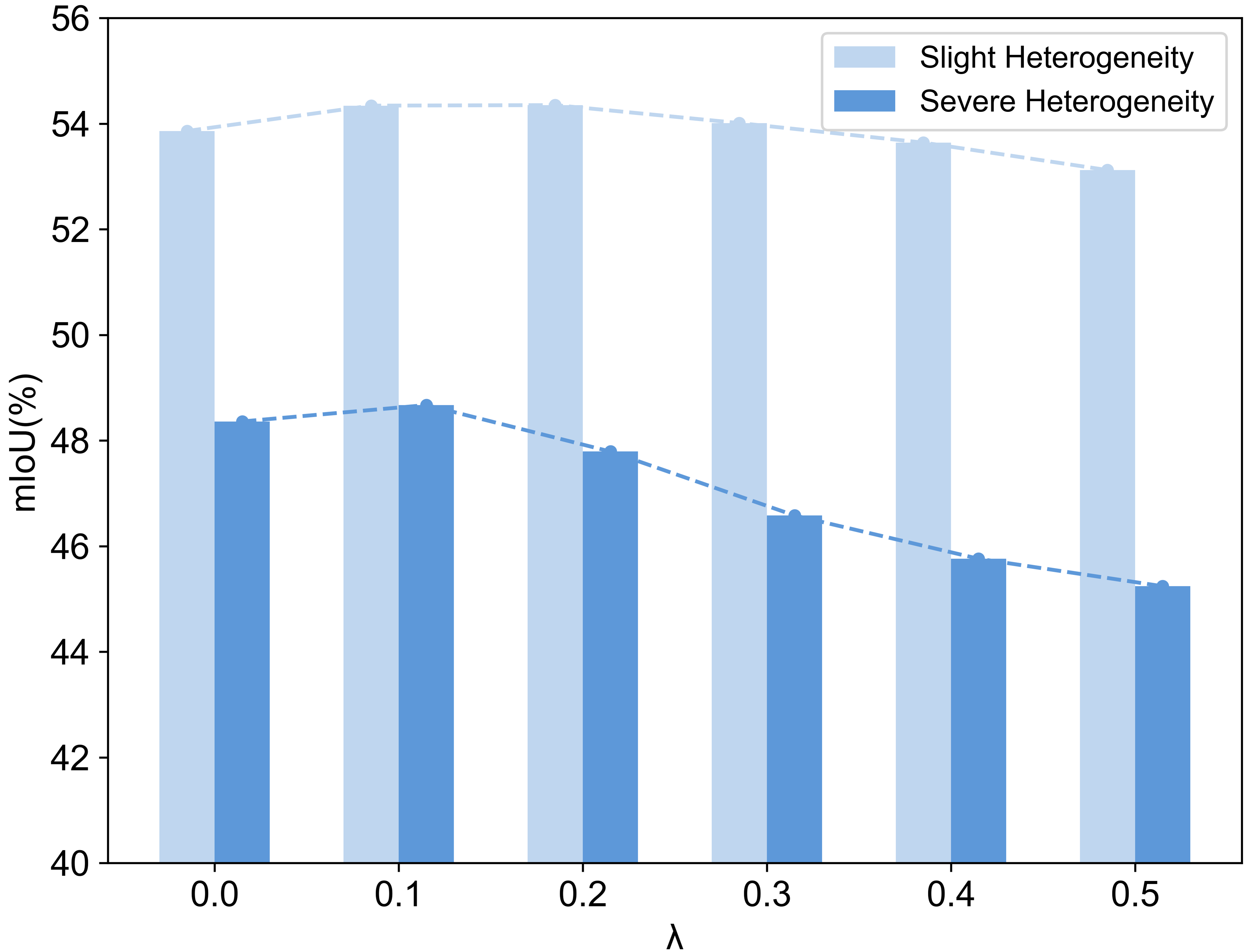}
\caption{Effect of the adversarial loss hyperparameter \(\lambda\).}
\label{fig3}
\end{figure}

\textbf{Effect of the Adversarial Loss Hyperparameter \(\lambda\)}.
Figure~\ref{fig3} examines the impact of the adversarial loss hyperparameter \(\lambda\) on performance under both slight and severe heterogeneity scenarios. It is evident that smaller values of \(\lambda\) improve performance, while further increases lead to a decline. This occurs because adversarial loss functions as an auxiliary constraint, and scaling it too high can negatively affect the segmentation loss. Additionally, fluctuations in \(\lambda\) are more pronounced under the strongly heterogeneous dataset, due to the greater disparity between global and local representations, which makes balancing both more challenging for the model. 

\begin{table}[ht]
\centering
\begin{tabular}{c|c|c}
\toprule
{Dataset} & {PSNR(dB)} & {SSIM(\%)} \\
\midrule
Slight Heterogeneity  & 14.52 & 36.73 \\
Severe Heterogeneity  & 11.46 & 29.01 \\
\bottomrule
\end{tabular}
\caption{Reconstruction results for class examples.} \label{table_4}
\end{table}

\begin{figure}[h]
\centering
\includegraphics [width=\columnwidth] {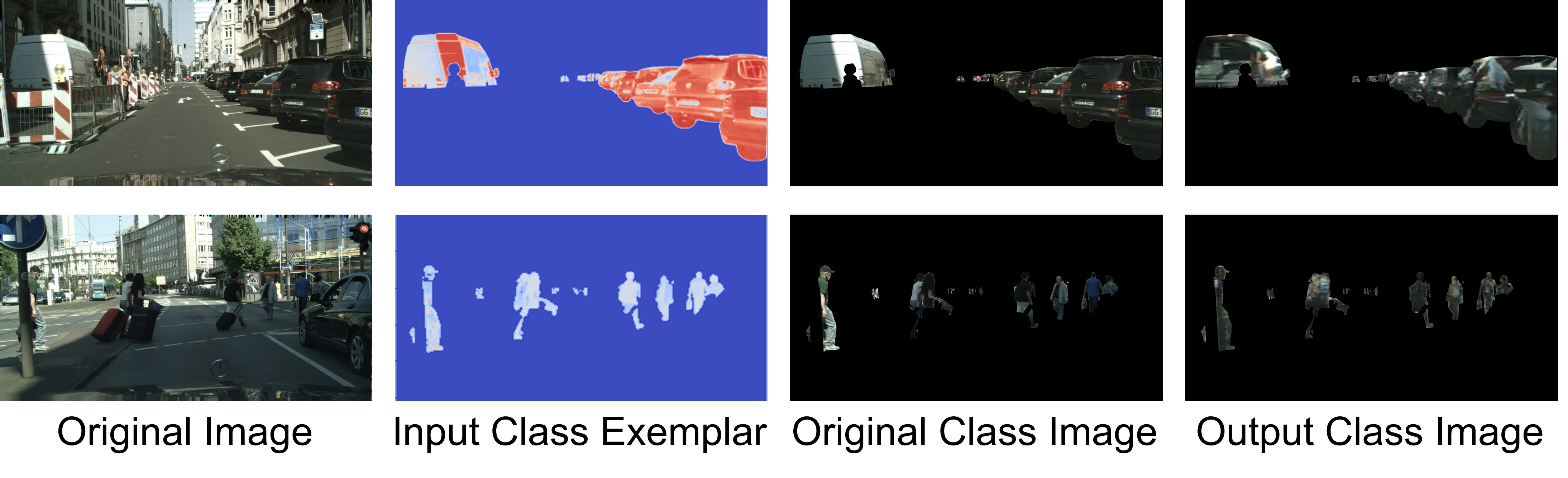}
\caption{Comparison of visual effects of original image and reconstruction attack.}
\label{fig4}
\end{figure}

\begin{table*}[!t]
\centering
\begin{tabular}{c|cccccccc}
\hline
\multirow{2}{*}{Methods} & \multicolumn{7}{c}{Category} \\
\cline{2-9}
& \textbf{Mean} & background & building & road & water & barren & forest & agriculture\\
\hline
FedAvg & 62.06 & 56.31 & 59.84 & 63.09 & 78.42 & 56.17 & 64.93 & 68.66 \\
GeoFed & 66.86 & 59.27 & 61.55 & 64.88 & 81.03 & \textbf{58.72} & 69.41 & \textbf{76.19} \\
FedSaaS & \textbf{68.22} & \textbf{60.38} & \textbf{63.07} & \textbf{64.90} & \textbf{84.24} & 57.53 & \textbf{70.85} & 75.96 \\
\hline
\end{tabular}
\caption{Comparison of mIoU scores (\%) for specific classes under LoveDA.} \label{table_4}
\end{table*}

\begin{table*}[!t]
\centering
\begin{tabular}{c|ccccccccc}
\hline
\multirow{2}{*}{Methods} & \multicolumn{8}{c}{Category} \\
\cline{2-10}
& \textbf{Mean} & Aorta & gallbladder & spleen & kidney & liver & pancreas & spleen & stomach\\
\hline
FedAvg & 68.21 & 82.45 & 66.97 & 61.03 & 63.82 & 77.15 & 42.70 & 81.31 & 69.02 \\
FedDP & 71.41 & 85.25 & 70.17 & 63.13 & 69.82 & 80.65 & 44.50 & \textbf{84.61} & 73.12\\
FedSaaS & \textbf{72.48} & \textbf{87.45} & \textbf{70.97} & \textbf{62.63} & \textbf{70.92} & \textbf{82.35} & \textbf{47.00} & 83.31 & \textbf{76.12}\\
\hline
\end{tabular}
\caption{Comparison of mIoU scores (\%) for specific classes under Synapse.} \label{table_5}
\end{table*}

\textbf{Privacy Leakage Risk Analysis}.
Uploading class exemplars may pose a risk of privacy leakage, as the uploaded information could potentially be stolen if the server is compromised. Therefore, we conduct image reconstruction experiments to illustrate the protection of FedSaaS for privacy data.

We use the method proposed by \cite{isola2017image} as the image reconstruction attack benchmark, which was trained on the Cityscapes dataset. This method has been proven to be effective and widely used by existing defense work \cite{wang2023privacy}. PSNR and SSIM are used to evaluate the effect of reconstructing the returned image through category examples. These two indicators are widely used quality assessment indicators in image and video processing, which are used to measure the quality difference between the reconstructed image or video and the original reference image. 

As shown in Table 4, PSNR and SSIM are less than 20 dB and 0.4 respectively, which is usually considered to be the result of reconstruction failure. At the same time, as shown in Figure~\ref{fig4}, the image generated by the reconstruction attack method has a great difference in high-level features from the original image and the corresponding category of the image, and no information related to the original image can be obtained from the generated image. We analyze three primary advantages of the class exemplars in resisting privacy attacks: 1) The class exemplars are highly sparse, lacking effective and informative features, which complicates the reconstruction process for potential attackers; 2) the storage order of the class exemplars on the server is random, hindering attackers' ability to infer the source of the data; 3) since the class exemplars represent highly abstracted feature representations, the absence of specific knowledge about the input network's type and parameters limits attackers' ability to reconstruct the original image effectively.

\begin{figure}[h]
	\centering
	\includegraphics [width=2.8in] {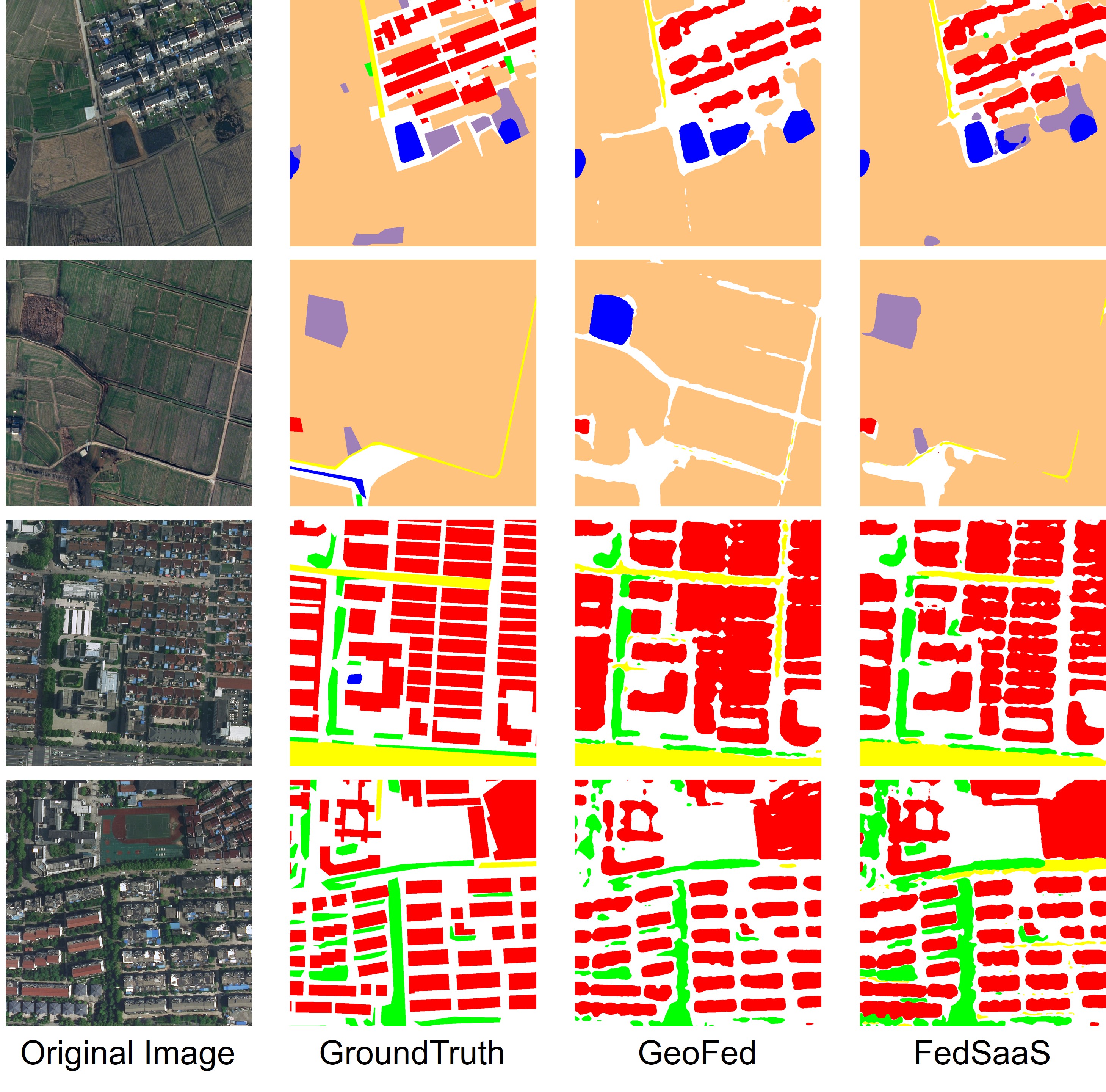}
	\caption{Comparison of visual results on LoveDA.}
	\label{figa5}
\end{figure}

\begin{figure}[h]
	\centering
	\includegraphics [width=2.8in] {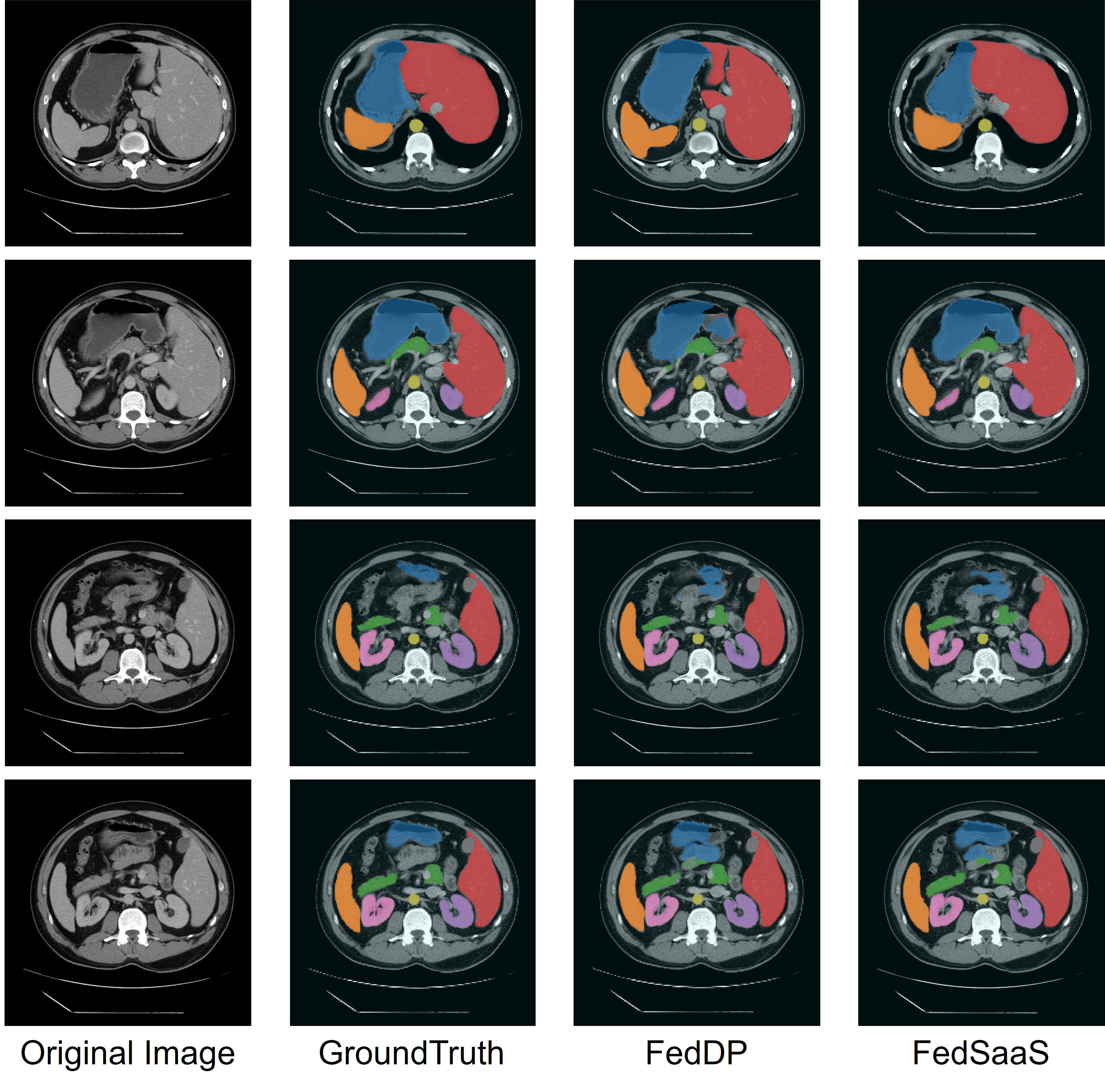}
	\caption{Comparison of visual results on Synapse.}
	\label{figa6}
\end{figure}

\section{Generalization Experiments}
To evaluate the performance of FedSaaS beyond autonomous driving scenarios, we conducted extended experiments in remote sensing image scenarios and multi-organ medical image scenarios to demonstrate the generalization ability.

\textbf{Remote Sensing image}.
For remote sensing image segmentation task, we evaluate on LoveDA \cite{junjue_wang_2021_5706578}, a widely adopted benchmark dataset for domain adaptation in remote sensing. Specifically designed to address urban-rural land cover segmentation challenges, LoveDA advances research in semantic segmentation and cross-domain transfer learning. The dataset comprises 5,987 high-resolution (0.3m) images with 166,768 meticulously annotated semantic objects, collected across three distinct cities. This diversity in geographic and urban-rural distributions enables rigorous testing of model generalizability under domain shifts. 

We divide the clients into urban/ rural groups, and follow the previous parameter settings for training and testing. We compare the two methods, fedavg and GeoFed \cite{tan2024bridging}, respectively. Table~\ref{table_4} shows the segmentation accuracy of each category. It can be seen that our method is almost leading in all aspects. Meanwhile, it can be seen from Figure~\ref{figa5} that GeoFed fails to identify the specified class in some areas, while FedSaaS is able to accurately locate and identify it accurately.

\textbf{Multi-organ medical image}.
For multi-organ medical image segmentation task, We use the Synapse dataset for evaluation, which is a publicly available multi-organ segmentation dataset from the MICCAI 2015 Multi-atlas Abdomen Labeling Challenge. It includes 30 abdominal CT scans, a total of 3779 axial enhanced abdominal clinical images, and the corresponding segmentation labels for 8 categories.

We divided the clients into 30 groups based on the scanned samples, and trained and tested them according to the previous parameter settings. We compared the two methods, fedavg and FedDP \cite{wang2023feddp}. Table~\ref{table_5} shows the segmentation accuracy of each category, and our method is in the leading position in the accuracy of each organ. Additionally, it can be seen from Figure~\ref{figa6} that FedDP has incorrect category recognition in some areas, while FedSaaS performs better in this regard.